\documentclass[sigconf]{acmart}

\usepackage{multirow}
\usepackage{amsmath}
\usepackage{enumitem}   % leftmargin
\setlist[itemize]{leftmargin=*,noitemsep,topsep=2pt,parsep=0pt}
\setlist[enumerate]{leftmargin=*,noitemsep,topsep=2pt,parsep=0pt}
\usepackage{algorithm}
\usepackage{algpseudocode}
\usepackage{graphicx}
\usepackage{booktabs,longtable}
\usepackage{booktabs,makecell}
\usepackage{microtype}
\usepackage{comment}
\usepackage{hyperref}
\usepackage{booktabs,subcaption,makecell}

\AtBeginDocument{%
  }

\copyrightyear{2026}
\acmYear{2026}
\setcopyright{cc}
\setcctype{by}
\acmConference[WSDM '26] {Proceedings of the Nineteenth ACM International Conference on Web Search and Data Mining}{February 22--26, 2026}{Boise, ID, USA.}
\acmBooktitle{Proceedings of the Nineteenth ACM International Conference on Web Search and Data Mining (WSDM '26), February 22--26, 2026, Boise, ID, USA}
\acmISBN{979-8-4007-2292-9/2026/02}
\acmDOI{10.1145/XXXXXX.XXXXXX}
\begin{document}

\title{Forget and Explain: Transparent Verification of GNN Unlearning}

\author{Imran Ahsan}

\affiliation{%
  \institution{Chung-Ang University}
  \department{Department of Smart Cities}
  \city{Seoul}
  \country{Republic of Korea}}
\email{imranahan23@cau.ac.kr}

\author{Hyunwook Yu}

\affiliation{%
  \institution{Chung-Ang University}
  \department{Department of Computer Science and Engineering}
  \city{Seoul}
  \country{Republic of Korea}
}
\email{yu990410@cau.ac.kr}

\author{Jinsung Kim}

\affiliation{%
  \institution{Chung-Ang University}
  \department{Department of Computer Science and Engineering}
  \city{Seoul}
  \country{Republic of Korea}
}
\email{kimjsung@cau.ac.kr}

\author{Mucheol Kim}
%\authornote{Corresponding author.}
%\orcid{0000-0002-7924-1186}
%\author{Jinsung Kim}
%\email{kimm@cau.ac.kr}
\affiliation{%
  \institution{Chung-Ang University}
  \department{Department of Computer Science and Engineering}
  \city{Seoul}
  \country{Republic of Korea}
}
\email{kimm@cau.ac.kr (corresponding)}
%\end{comment}

\renewcommand{\shortauthors}{Imran et al.}

\begin{abstract}
  Graph neural networks (GNNs) are increasingly used to model complex patterns in graph-structured data. However, enabling them to ``forget'' designated information remains challenging, especially under privacy regulations such as the GDPR. Existing unlearning methods largely optimize for efficiency and scalability, yet they offer little transparency, and the black-box nature of GNNs makes it difficult to verify whether forgetting has truly occurred. We propose an \emph{explainability-driven verifier} for GNN unlearning that snapshots the model before and verifies after deletion, using attribution shifts and localized structural changes (e.g., graph edit distance) as transparent evidence. The verifier uses five explainability metrics—residual attribution, heatmap shift, explainability score deviation, graph edit distance, and (diagnostic) graph rule shift. We evaluate two backbones (GCN, GAT) and four unlearning strategies (Retrain, GraphEditor, GNNDelete, IDEA) across five benchmarks (Cora, Citeseer, Pubmed, Coauthor‑CS, Coauthor‑Physics). Results show that Retrain and GNNDelete achieve near-complete forgetting, GraphEditor provides partial erasure, and IDEA leaves residual signals. These explanation deltas provide the primary, human-readable evidence of forgetting; we also report membership‑inference ROC‑AUC as a complementary, graph‑wide privacy signal. The code is available at 
  \url{https://github.com/ImranAhsan23/F-E}
  %\url{https://anonymous.4open.science/r/TransparentGNNUnlearning}
\end{abstract}



%Data Erasure Compliance;


\maketitle

\section{Introduction}
In graph neural networks (GNNs), message passing propagates signals across $k$-hop neighborhoods, so removing a single item can leave information traces nearby. This elevates verifiable forgetting to a core requirement in privacy-sensitive settings—for example, the right to erasure under the General Data Protection Regulation (GDPR)~\cite{R9}. GNNs have emerged as powerful models for capturing complex dependencies in graph-structured data~\cite{R1}, with demonstrated effectiveness in traffic prediction~\cite{R2}, drug discovery~\cite{R3}, healthcare~\cite{R4}, and recommendation systems~\cite{R5}. Unlearning refers to removing the influence of specific nodes, edges, or subgraphs so that the model behaves as if those data were absent from training, motivating explicit, auditable verification of forgetting~\cite{R6}.

However, existing approaches primarily optimize for efficiency and scalability, but often neglect transparency, leaving unresolved concerns about trust, privacy compliance, and reliable verification of unlearning outcomes~\cite{R7}.
Two key challenges stand out.
First, GDPR grants users the right to request the removal of their data from a model~\cite{R9}.
Deterministic unlearning techniques address this by retraining the model from scratch to guarantee complete removal~\cite{R10,R11,R12}, while approximate approaches offer faster updates with bounded residual influence, including those tailored for GNNs~\cite{R8,R13,R14}.
Second, the inherent opacity of GNNs complicates verification, motivating Explainable-AI (XAI) studies that aim to identify influential nodes, edges, and subgraphs~\cite{R15,R16,R17,R18}.
Existing approaches fail to quantify how thoroughly a GNN forgets designated graph components and provide little to no human-readable evidence of the forgetting process.

To address this gap, we propose a practical, explainability-driven verification framework for GNN unlearning.

\begin{figure*}[h]
  \centering
  \includegraphics[width=\linewidth]{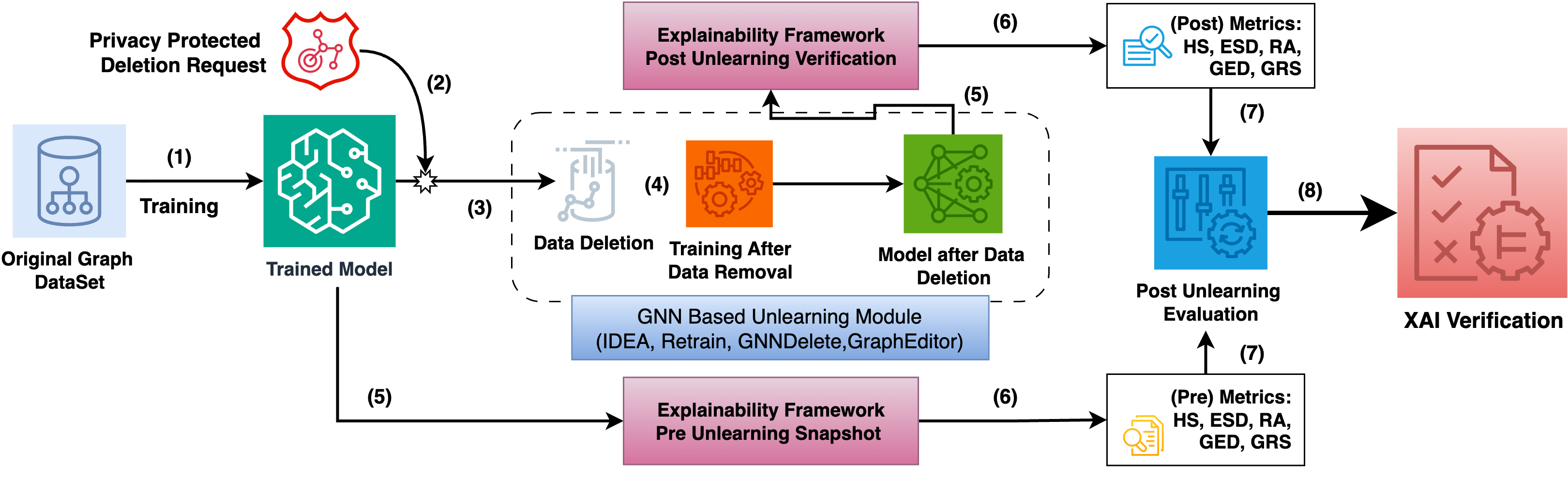}
  \caption{Overview of the explainability-driven verification pipeline 
  } \label{fig:fig1}
  \Description{This figure illustrates the main architecture of this proposed approach, which elaborates the step by step procedure of the working of our proposed method.}
\end{figure*}

\begin{itemize}[leftmargin=*]
    \item \textbf{Unified pipeline.} 
    We integrate GraphChef~\cite{R16} and ProxyGraph-inspired~\cite{R18} k-hop local proxies with attribution heatmaps to trace information flow before and after data deletion.
    \item \textbf{New metrics.} 
    We introduce a suite of attribution- and structure-based metrics such as  Residual Attribution (RA), Heatmap Shift (HS), Explainability Score Deviation (ESD), Graph Edit Distance (GED), and Graph Rule Shift (GRS), to provide fine-grained, quantitative evidence of forgetting.
    \item \textbf{Empirical validation.} Benchmark results show residual memorization that accuracy and Membership-Inference (MI) tests overlook. 
    Our framework addresses this gap by verifying, tracking, and explaining the unlearning process in existing GNN methods.
\end{itemize}

\section{Related Work}
Existing unlearning pipelines typically evaluate success using indirect signals such as changes in accuracy or membership-inference risk~\cite{R27}.
However, their metrics cannot reveal specific regions where a model may still retain information from deleted subgraphs. 
At the same time, XAI studies rarely examine the impact of graph unlearning. 
A few works explore vision models using measures such as Heatmap Coverage and Attention Shift to verify forgetting~\cite{R19}, while previous research applies saliency primarily to detect backdoor triggers in GNNs, leaving its potential for unlearning verification unexplored~\cite{R28}.

\section{Methodology}
We begin by training a baseline GNN model on the input graph dataset, as shown in Figure~\ref{fig:fig1}.
When a GDPR-compliant deletion request is received, an unlearning module---Retrain, GNNDelete~\cite{R11}, GraphEditor~\cite{R30}, or IDEA~\cite{R12}---removes the designated nodes, yielding a model-after-deletion through retraining.
Our explain-ability-driven framework surrounds this unlearning step.
It first generates attribution heatmaps, then computes the metrics in Section~\ref{subsec:metrics}, including auxiliary GRS.
A second round of metrics is obtained by applying the same explanation procedures to the updated model.
A comparison of these vectors with the baseline reveals whether forgetting is complete, partial, or ineffective. It yields a concise XAI evidence profile for regulators, developers, and affected users.
The subsections below detail each stage of the framework.

\noindent\textbf{Notation.}
$G=(V,E,\mathbf{X})$ denotes the original attributed graph,  
$|V|=n$, feature matrix $\mathbf{X}\!\in\!\mathbb{R}^{n\times d}$,
and $f_{\boldsymbol{\theta}}$ the trained GNN.
The nodes to be forgotten are $F\subseteq V$, $|F|=m$.
After unlearning we obtain a new model $f_{\boldsymbol{\theta}'}$
and a modified graph $G'=(V',E',\mathbf{X}')$.

\subsection{Metrics}
\label{subsec:metrics}
This section introduces five metrics—RA, HS, ESD, GED, and GRS—that capture attribution-, structural-, and rule-level changes and offer a transparent basis for benchmarking unlearning. 
GED quantifies structural deviation after node removal, HS measures attribution divergence~\cite{R25}, RA aligns attribution with predictions~\cite{R23}, and ESD reflects changes in feature importance~\cite{R24}. 
GRS is reported only as a diagnostic, since surrogate trees may vary independently of forgetting. 
Formal definitions are provided to ensure reproducibility. 
In addition, we track membership-inference using the standard loss‑threshold attack and report ROC‑AUC between members and non‑members; lower post‑unlearning AUC indicates reduced leakage.
RA captures direct influence through $F$ (0 when targets are removed), while HS/ESD/GED$\Delta$ quantify indirect neighborhood or structural effects. 
GRS serves only as a diagnostic surrogate.

\noindent\textit{\textbf{(1) Residual Attribution (RA).}}
The proportion of the total attribution that continues to propagate through the forgotten nodes.
\[
   \mathrm{RA}
   \;=\;
   (\displaystyle\sum_{v\in F} a^{\text{post}}_{v} / \displaystyle\sum_{u\in V'}a^{\text{post}}_{u})
   % \frac{\displaystyle\sum_{v\in F} a^{\text{post}}_{v}}
   %      {\displaystyle\sum_{u\in V'} a^{\text{post}}_{u}}
   \times 100\%.
\]

\noindent\textit{\textbf{(2) Heatmap Shift (HS).}}
The mean absolute change in attribution across nodes.
\[
   \mathrm{HS}
   \;=\;
   \frac{1}{n}\sum_{v\in V} \bigl|\,a^{\text{pre}}_{v}-a^{\text{post}}_{v}\bigr|.
\]

\noindent\textit{\textbf{(3) Explainability Score Deviation (ESD).}}
A measure of the variation in feature importance.
\[
   \mathrm{ESD}
   \;=\;
   \frac{1}{|F|}\sum_{v\in F}
   \bigl|\,a^{\text{pre}}_{v}-a^{\text{post}}_{v}\bigr|.
\]

\noindent\textit{\textbf{(4) Graph Edit Distance (GED).}}
It quantifies the difference in the number of edges between the two proxy graphs.
All GED values are computed on k‑hop local proxies (ProxyGraph‑inspired) extracted with a fixed configuration; consequently, we report GED as a relative indicator of structural change within a dataset, rather than as an absolute measure across datasets.
\[
   \mathrm{GED}
   \;=\;
   \bigl|\,E\triangle E'\bigr|,
   \quad
   E\triangle E'=(E\setminus E')\cup(E'\setminus E),
\]

\noindent\textit{\textbf{(5) Graph Rule Shift (GRS).}}
The number of decision rules that disappear following the unlearning process.
\[
   \mathrm{GRS}
   \;=\;
   \bigl|\mathcal{R}^{\text{pre}}\bigr|
   \;-\;
   \bigl|\mathcal{R}^{\text{post}}\bigr|.
\]

\subsection{Pre Unlearning Explanation Snapshot}
This stage identifies how the model depends on the data targeted for removal by first locating the relevant nodes.
Explainability methods probe the trained model to produce attribution heatmaps, local proxy subgraphs, and surrogate rules that reveal how target entities affect predictions. These artifacts are archived as the baseline snapshot and used to compute influence scores that quantify each target’s contribution.
\noindent\textbf{Gradient heatmap}
The approach follows the vanilla saliency map of Simonyan et al.,\cite{R29} to compute a saliency value for every node $v\!\in\!V$
\begin{equation}
        a^{\text{pre}}_{v}
        \;=\;
        \bigl\lVert
          \nabla_{\mathbf{x}_{v}}\,
          \ell\!\bigl(f_{\boldsymbol{\theta}}(\mathbf{X},E),y\bigr)
        \bigr\rVert_{1},
        \label{eq:grad}
\end{equation}
where $\ell$ is cross-entropy w.r.t.\ a mini-batch of labels~$y$.
The vector
$\mathbf{a}^{\text{pre}}=[a^{\text{pre}}_{1},\dots,a^{\text{pre}}_{n}]$
forms the baseline heatmap. 
We adopt vanilla saliency as a reference explainer for its implementation simplicity and reproducibility; 
because we analyze within‑run pre $\rightarrow$ post deltas and report per‑dataset contrasts, our use does not hinge on absolute gradient scale---mitigating common saturation/scale effects.
\noindent\textbf{Local proxy graph.}
This technique follows the approach inspired by ProxyGraph \cite{R18}, which constructs localized structural representations around target nodes. For each node $v\!\in\!F$, the method extracts a k-hop ego-network, capturing the local neighborhood. The union of all such subgraphs defines the local proxy graph $\hat{G}^{\text{pre}}=(\hat{V},\hat{E})$. This graph serves as a structural fingerprint of the region selected for removal and constitutes the baseline proxy prior to unlearning.
\noindent\textbf{Symbolic rule set.}
The approach leverages GraphChef \cite{R16}, which distills the black-box GNN $f_{\boldsymbol{\theta}}$ into an interpretable surrogate model. Specifically, a depth-controlled decision tree fits the model’s predictions. The full set of root-to-leaf paths defines the symbolic rule set $\mathcal{R}^{\text{pre}}$. Each path in $\mathcal{R}^{\text{pre}}$ captures a logical predicate that approximates the model's prediction mechanism. This forms a rule-based baseline explanation prior to unlearning.

\begin{table*}[ht]
\centering
\scriptsize
\setlength{\tabcolsep}{2pt}
\renewcommand{\arraystretch}{0.95}
\begin{tabular}{l l
                r r r r r r
                r r r r r r}
\toprule

& & \multicolumn{6}{c}{\textbf{GCN}} & \multicolumn{6}{c}{\textbf{GAT}}\\
\cmidrule(lr){3-8}\cmidrule(lr){9-14}
Dataset & Method
& RA$_\mathrm{pre}$ (\%) & RA$_\mathrm{post}$ (\%) & HS $\uparrow$ & ESD $\uparrow$  & GED$\Delta$ $\uparrow$ & GRS
& RA$_\mathrm{pre}$ (\%) & RA$_\mathrm{post}$ (\%) &  HS $\uparrow$ & ESD $\uparrow$  & GED$\Delta$ $\uparrow$ & GRS \\
\midrule
Cora & Retrain & 4.77\% $\pm$ 1.07 & 0.00\% $\pm$ 0.00 & 0.016$\pm$0.001 & 0.014$\pm$0.001 & 6,975$\pm$50 & 4.67$\pm$1.53 & 5.27\% $\pm$ 0.20 & 0.00\% $\pm$ 0.00 & 0.031$\pm$0.003 & 0.031$\pm$0.003 & 7,151$\pm$21 & 4.67$\pm$2.08 \\
 & GNNDelete & 4.74\% $\pm$ 0.99 & 0.00\% $\pm$ 0.00 & 0.015$\pm$0.001 & 0.014$\pm$0.002 & 6,975$\pm$50 & 2.00$\pm$2.00 & 5.28\% $\pm$ 0.12 & 0.00\% $\pm$ 0.00 & 0.032$\pm$0.001 & 0.032$\pm$0.003 & 7,151$\pm$21 & 3.67$\pm$0.58 \\
 & GraphEditor & 4.70\% $\pm$ 1.14 & 5.32\% $\pm$ 1.83 & 0.006$\pm$0.000 & 0.012$\pm$0.003 & 1,113$\pm$48 & 7.33$\pm$0.58 & 5.11\% $\pm$ 0.30 & 3.95\% $\pm$ 0.31 & 0.010$\pm$0.001 & 0.022$\pm$0.002 & 1,134$\pm$107 & 7.33$\pm$0.58 \\
 & IDEA & 5.00\% $\pm$ 0.20 & 4.77\% $\pm$ 1.16 & 0.005$\pm$0.001 & 0.005$\pm$0.001 & 0$\pm$0 & 1.67$\pm$2.89 & 5.20\% $\pm$ 0.27 & 5.10\% $\pm$ 0.20 & 0.005$\pm$0.001 & 0.005$\pm$0.001 & 0$\pm$0 & 2.67$\pm$2.52 \\
% \midrule
Citeseer & Retrain & 5.60\% $\pm$ 1.02 & 0.00\% $\pm$ 0.00 & 0.028$\pm$0.001 & 0.029$\pm$0.003 & 4,803$\pm$78 & 4.00$\pm$1.00 & 4.61\% $\pm$ 0.33 & 0.00\% $\pm$ 0.00 & 0.024$\pm$0.001 & 0.024$\pm$0.001 & 4,611$\pm$48 & 3.67$\pm$0.58 \\
 & GNNDelete & 5.23\% $\pm$ 0.64 & 0.00\% $\pm$ 0.00 & 0.027$\pm$0.001 & 0.028$\pm$0.002 & 4,803$\pm$78 & 3.67$\pm$1.15 & 4.63\% $\pm$ 0.26 & 0.00\% $\pm$ 0.00 & 0.023$\pm$0.001 & 0.023$\pm$0.001 & 4,611$\pm$48 & 4.00$\pm$1.00 \\
 & GraphEditor & 5.56\% $\pm$ 0.86 & 5.17\% $\pm$ 2.04 & 0.012$\pm$0.001 & 0.022$\pm$0.003 & 962$\pm$48 & 8.00$\pm$0.00 & 4.58\% $\pm$ 0.31 & 4.01\% $\pm$ 0.37 & 0.011$\pm$0.001 & 0.023$\pm$0.004 & 853$\pm$132 & 8.00$\pm$0.00 \\
 & IDEA & 4.97\% $\pm$ 0.21 & 4.95\% $\pm$ 0.21 & 0.012$\pm$0.001 & 0.012$\pm$0.003 & 0$\pm$0 & 3.00$\pm$1.00 & 4.70\% $\pm$ 0.33 & 4.76\% $\pm$ 0.15 & 0.010$\pm$0.001 & 0.009$\pm$0.000 & 0$\pm$0 & 2.00$\pm$1.00 \\
% \midrule
Pubmed & Retrain & 4.86\% $\pm$ 0.11 & 0.00\% $\pm$ 0.00 & 0.023$\pm$0.000 & 0.020$\pm$0.000 & 72,780$\pm$154 & 6.00$\pm$2.00 & 4.65\% $\pm$ 0.22 & 0.00\% $\pm$ 0.00 & 0.022$\pm$0.000 & 0.020$\pm$0.000 & 72,573$\pm$123 & 5.00$\pm$2.00 \\
 & GNNDelete & 4.86\% $\pm$ 0.11 & 0.00\% $\pm$ 0.00 & 0.023$\pm$0.000 & 0.020$\pm$0.000 & 72,780$\pm$154 & 2.00$\pm$0.00 & 4.67\% $\pm$ 0.23 & 0.00\% $\pm$ 0.00 & 0.022$\pm$0.000 & 0.020$\pm$0.000 & 72,573$\pm$123 & 2.00$\pm$0.00 \\
 & GraphEditor & 4.23\% $\pm$ 0.29 & 5.03\% $\pm$ 0.29 & 0.015$\pm$0.000 & 0.015$\pm$0.000 & 11,066$\pm$113 & 8.00$\pm$0.00 & 4.76\% $\pm$ 0.20 & 4.83\% $\pm$ 0.15 & 0.002$\pm$0.000 & 0.002$\pm$0.000 & 11,286$\pm$120 & 8.00$\pm$0.00 \\
 & IDEA & 4.99\% $\pm$ 0.17 & 4.96\% $\pm$ 0.15 & 0.001$\pm$0.000 & 0.001$\pm$0.000 & 0$\pm$0 & 4.67$\pm$1.15 & 4.67\% $\pm$ 0.15 & 4.64\% $\pm$ 0.12 & 0.001$\pm$0.000 & 0.001$\pm$0.000 & 0$\pm$0 & 4.33$\pm$0.58 \\
% \midrule
Coauthor & Retrain & 4.66\% $\pm$ 0.24 & 0.00\% $\pm$ 0.00 & 0.004$\pm$0.000 & 0.004$\pm$0.000 & 144,458$\pm$237 & 0.67$\pm$1.15 & 4.93\% $\pm$ 0.22 & 0.00\% $\pm$ 0.00 & 0.003$\pm$0.000 & 0.003$\pm$0.000 & 144,622$\pm$262 & 4.33$\pm$1.15 \\
(CS) & GNNDelete & 4.61\% $\pm$ 0.24 & 0.00\% $\pm$ 0.00 & 0.004$\pm$0.000 & 0.004$\pm$0.000 & 144,458$\pm$237 & 4.66$\pm$1.10 & 4.91\% $\pm$ 0.25 & 0.00\% $\pm$ 0.00 & 0.010$\pm$0.000 & 0.010$\pm$0.000 & 144,622$\pm$109 & 4.67$\pm$1.15 \\
 & GraphEditor & 4.96\% $\pm$ 0.09 & 4.70\% $\pm$ 0.20 & 0.002$\pm$0.000 & 0.003$\pm$0.000 & 15,632$\pm$621 & 8.00$\pm$0.00 & 5.14\% $\pm$ 0.20 & 3.80\% $\pm$ 0.44 & 0.003$\pm$0.000 & 0.005$\pm$0.000 & 15,551$\pm$672 & 2.00$\pm$0.00 \\
 & IDEA & 4.97\% $\pm$ 0.14 & 4.85\% $\pm$ 0.14 & 0.001$\pm$0.000 & 0.001$\pm$0.000 & 0$\pm$0 & 3.00$\pm$5.00 & 4.93\% $\pm$ 0.13 & 4.95\% $\pm$ 0.15 & 0.002$\pm$0.000 & 0.002$\pm$0.000 & 0$\pm$0 & 1.00$\pm$0.00 \\
% \midrule
Coauthor & Retrain & 4.83\% $\pm$ 0.43 & 0.00\% $\pm$ 0.00 & 0.005$\pm$0.000 & 0.004$\pm$0.000 & 456,456$\pm$286 & 3.00$\pm$2.00 & 5.08\% $\pm$ 0.35 & 0.00\% $\pm$ 0.00 & 0.005$\pm$0.000 & 0.004$\pm$0.000 & 455,743$\pm$306 & 2.00$\pm$1.00 \\
(Physics) & GNNDelete & 4.75\% $\pm$ 0.37 & 0.00\% $\pm$ 0.00 & 0.005$\pm$0.000 & 0.004$\pm$0.000 & 456,456$\pm$286 & 2.00$\pm$0.00 & 5.00\% $\pm$ 0.38 & 0.00\% $\pm$ 0.00 & 0.005$\pm$0.000 & 0.004$\pm$0.000 & 455,743$\pm$306 & 2.00$\pm$1.00 \\
 & GraphEditor & 4.93\% $\pm$ 0.12 & 4.54\% $\pm$ 0.41 & 0.002$\pm$0.000 & 0.003$\pm$0.000 & 30,480$\pm$302 & 8.00$\pm$0.00 & 5.22\% $\pm$ 0.17 & 3.95\% $\pm$ 0.04 & 0.003$\pm$0.000 & 0.004$\pm$0.000 & 30,923$\pm$336 & 8.00$\pm$0.00 \\
 & IDEA & 4.99\% $\pm$ 0.21 & 4.96\% $\pm$ 1.16 & 0.001$\pm$0.000 & 0.001$\pm$0.000 & 0$\pm$0 & 2.67$\pm$0.00 & 4.90\% $\pm$ 0.15 & 4.98\% $\pm$ 0.20 & 0.002$\pm$0.000 & 0.002$\pm$0.000 & 0$\pm$0 & 4.00$\pm$0.00 \\
\bottomrule
\end{tabular}
\caption{Explainability metrics \emph{before} and \emph{after} unlearning. Columns: $RA_{\text{pre}}$, $RA_{\text{post}}$; HS and ESD are pre$\to$post attribution shifts; GED$\Delta$ is the edit distance between pre/post proxy graphs; GRS is the number of rules removed after unlearning. Left block: GCN; right block: GAT. Values are mean$\pm$std over seeds.}
\label{tab:exp_metrics_gcn_gat}
\end{table*}

\subsection{Post Unlearning Verification}
Post-unlearning verification evaluates the effectiveness and interpretability of forgetting. 
It reapplies pre-unlearning XAI methods to the updated model to generate new attribution maps, then compares them with the baseline using five metrics: (1) HS captures attribution shifts, (2) GED detects structural variations in explanation subgraphs, (3) ESD reflects changes in fidelity and sparsity, (4) RA measures residual influence from the forgotten entity, and (5) GRS quantifies shifts in symbolic rule sets.
This end-to-end process evaluates unlearning evidence while preserving verifiability, interpretability, and trustworthiness for developers and users.

\begin{table*}[t]
\centering
\scriptsize
\setlength{\tabcolsep}{2pt}
\renewcommand{\arraystretch}{0.95}
\begin{tabular}{l l r c r c}
\toprule
 & & \multicolumn{2}{c}{\textbf{GCN}} & \multicolumn{2}{c}{\textbf{GAT}}\\
\cmidrule(lr){3-4}\cmidrule(lr){5-6}
Dataset & Method & $\Delta$RA (\%)$\uparrow$ & MI ROC-AUC (Pre$\rightarrow$Post)$\downarrow$ & $\Delta$RA (\%)$\uparrow$ & MI ROC-AUC (Pre$\rightarrow$Post)$\downarrow$\\
\midrule
Cora        & Retrain & 4.77\% $\pm$ 1.07 & 0.520$\pm$0.017\,$\rightarrow$\,0.525$\pm$0.010 & 5.27\% $\pm$ 0.20 & 0.520$\pm$0.023\,$\rightarrow$\,0.511$\pm$0.017 \\
            & GNNDelete & 4.74\% $\pm$ 0.99 & 0.520$\pm$0.017\,$\rightarrow$\,0.515$\pm$0.006 & 5.28\% $\pm$ 0.12 & 0.520$\pm$0.022\,$\rightarrow$\,0.506$\pm$0.003 \\
            & GraphEditor & -0.62\% $\pm$ 1.83 & 0.520$\pm$0.017\,$\rightarrow$\,0.521$\pm$0.007 & 1.16\% $\pm$ 0.31 & 0.520$\pm$0.019\,$\rightarrow$\,0.513$\pm$0.017 \\
            & IDEA & 0.23\% $\pm$ 0.20 & 0.519$\pm$0.018\,$\rightarrow$\,0.519$\pm$0.014 & 0.10\% $\pm$ 0.20 & 0.519$\pm$0.018\,$\rightarrow$\,0.519$\pm$0.014 \\
Citeseer    & Retrain & 5.60\% $\pm$ 1.02 & 0.559$\pm$0.012\,$\rightarrow$\,0.553$\pm$0.004 & 4.61\% $\pm$ 0.33 & 0.550$\pm$0.016\,$\rightarrow$\,0.553$\pm$0.008 \\
            & GNNDelete & 5.23\% $\pm$ 0.64 & 0.559$\pm$0.012\,$\rightarrow$\,0.526$\pm$0.001 & 4.63\% $\pm$ 0.26 & 0.550$\pm$0.015\,$\rightarrow$\,0.536$\pm$0.007 \\
            & GraphEditor & 0.38\% $\pm$ 1.83 & 0.559$\pm$0.012\,$\rightarrow$\,0.561$\pm$0.007 & 0.57\% $\pm$ 0.31 & 0.550$\pm$0.015\,$\rightarrow$\,0.556$\pm$0.015 \\
            & IDEA & 0.01\% $\pm$ 0.15 & 0.559$\pm$0.012\,$\rightarrow$\,0.557$\pm$0.012 & -0.06\% $\pm$ 0.11 & 0.550$\pm$0.017\,$\rightarrow$\,0.547$\pm$0.014 \\
Pubmed      & Retrain & 4.86\% $\pm$ 0.11 & 0.509$\pm$0.006\,$\rightarrow$\,0.512$\pm$0.007 & 4.65\% $\pm$ 0.22 & 0.502$\pm$0.004\,$\rightarrow$\,0.504$\pm$0.005 \\
            & GNNDelete & 4.86\% $\pm$ 0.11 & 0.509$\pm$0.006\,$\rightarrow$\,0.509$\pm$0.005 & 4.67\% $\pm$ 0.23 & 0.502$\pm$0.004\,$\rightarrow$\,0.503$\pm$0.005 \\
            & GraphEditor & -0.80\% $\pm$ 0.16 & 0.509$\pm$0.006\,$\rightarrow$\,0.518$\pm$0.004 & 0.07\% $\pm$ 0.20 & 0.505$\pm$0.003\,$\rightarrow$\,0.508$\pm$0.004 \\
            & IDEA & 0.03\% $\pm$ 0.10 & 0.509$\pm$0.006\,$\rightarrow$\,0.506$\pm$0.004 & 0.03\% $\pm$ 0.10 & 0.509$\pm$0.006\,$\rightarrow$\,0.506$\pm$0.005 \\
Coauthor    & Retrain & 4.66\% $\pm$ 0.24 & 0.511$\pm$0.001\,$\rightarrow$\,0.511$\pm$0.006 & 4.93\% $\pm$ 0.22 & 0.509$\pm$0.002\,$\rightarrow$\,0.512$\pm$0.004 \\
(CS)        & GNNDelete & 4.61\% $\pm$ 0.24 & 0.511$\pm$0.001\,$\rightarrow$\,0.508$\pm$0.006 & 4.91\% $\pm$ 0.25 & 0.511$\pm$0.001\,$\rightarrow$\,0.508$\pm$0.005 \\
            & GraphEditor & -0.25\% $\pm$ 0.36 & 0.510$\pm$0.001\,$\rightarrow$\,0.513$\pm$0.001 & 1.34\% $\pm$ 0.30 & 0.509$\pm$0.001\,$\rightarrow$\,0.509$\pm$0.002 \\
            & IDEA & -0.12\% $\pm$ 0.14 & 0.510$\pm$0.001\,$\rightarrow$\,0.508$\pm$0.001 & -0.02\% $\pm$ 0.13 & 0.509$\pm$0.002\,$\rightarrow$\,0.507$\pm$0.003 \\
Coauthor    & Retrain & 4.83\% $\pm$ 0.43 & 0.503$\pm$0.001\,$\rightarrow$\,0.507$\pm$0.002 & 5.08\% $\pm$ 0.35 & 0.501$\pm$0.002\,$\rightarrow$\,0.504$\pm$0.003 \\
(Physics)   & GNNDelete & 4.75\% $\pm$ 0.37 & 0.503$\pm$0.001\,$\rightarrow$\,0.505$\pm$0.002 & 5.00\% $\pm$ 0.38 & 0.503$\pm$0.001\,$\rightarrow$\,0.505$\pm$0.002 \\
            & GraphEditor & 0.39\% $\pm$ 0.35 & 0.503$\pm$0.001\,$\rightarrow$\,0.503$\pm$0.003 & 1.27\% $\pm$ 0.17 & 0.503$\pm$0.001\,$\rightarrow$\,0.503$\pm$0.001 
\\
            & IDEA & 0.03\% $\pm$ 0.18 & 0.502$\pm$0.001\,$\rightarrow$\,0.502$\pm$0.001 & -0.08\% $\pm$ 0.12 & 0.502$\pm$0.001\,$\rightarrow$\,0.503$\pm$0.001 \\
\bottomrule
\end{tabular}
\caption{Residual attribution change ($\Delta$RA) and \textbf{MI ROC-AUC} before$\to$after unlearning for all datasets and both backbones. Lower post MI AUC indicates lower membership leakage; larger $\Delta$RA indicates stronger forgetting. 
}
\label{tab:ra_mi_gcn_gat}
\end{table*}

\section{Experimental Setup}
The evaluation addresses two conceptual dimensions of unlearning: transparency and effectiveness. 
\textbf{RQ1:} Does the unlearning process visibly remove the targeted information?
\textbf{RQ2:} Is the model no longer influenced by the unlearned data?

\noindent\textbf{Datasets}
We evaluate the method on canonical benchmarks including Cora, Citeseer, Pubmed, and the Coauthor‑CS/Coauthor‑Physics graphs, widely used in the GNN and unlearning literature for transductive node classification. 
All experiments adopt the standard fixed train/val/test splits, kept identical across methods and backbones.

\noindent\textbf{Implementation Details}
The experimental configuration employs a two-layer GCN/GAT backbone (64 hidden units, ReLU, dropout 0.5) trained with Adam (lr=0.005) for 100 epochs, and evaluates Retrain, GNNDelete, GraphEditor, and IDEA. For each dataset, the forget set $F$ contains $5\%$ of nodes sampled uniformly without replacement; we adopt $5\%$ to reflect an individual-level deletion budget while preserving graph connectivity and stable attribution/MI estimates. For a given seed, the sampled $F$ is held fixed across all methods and both backbones (different seeds induce independent $F$ samples). The local ProxyGraph is the union of \textbf{$k{=}2$}-hop ego networks around $F$ (undirected; original attributes preserved, no relabeling); $\mathrm{GED}\Delta$ is computed as the edge symmetric difference between the pre/post proxies. We run three seeds $\{1001,1002,1003\}$ and report mean$\pm$std. Reported metrics are RA, HS, ESD, $\mathrm{GED}\Delta$, GRS (GraphChef depth‑3 surrogate), and MI ROC‑AUC (Pre$\to$Post).

\section{Results and Analysis}\label{sec:results}
This section presents a comprehensive account of the evaluation outcomes. Tables~\ref{tab:exp_metrics_gcn_gat} and~\ref{tab:ra_mi_gcn_gat} summarize our evaluation results. 
Across datasets and both backbones, Retrain and GNNDelete deliver strong forgetting---$\mathrm{RA}_{\text{post}} \approx 0$ (as expected for deletion-based methods that remove the designated targets; see Section~\ref{subsec:metrics}), HS/ESD show substantive pre$\to$post attribution shifts, and the local proxy exhibits a large $\mathrm{GED}\Delta$---while GraphEditor yields partial erasure ($\mathrm{GED}\Delta>0$ but smaller than retraining methods) with $\Delta\mathrm{RA}$ small or occasionally negative (a ratio effect in which pruning neighbors concentrates attribution on retained targets even as absolute saliency falls), aligning with residual attribution and indicating that local edits reduce but do not fully eliminate influence pathways; by contrast, IDEA remains essentially unchanged ($\mathrm{GED}\Delta \approx 0$; $\mathrm{RA}_{\text{post}} \approx \mathrm{RA}_{\text{pre}}$). Because HS/ESD inherit gradient scale, we interpret attribution shifts strictly within-dataset and avoid cross-dataset magnitude comparisons (see Section~\ref{subsec:metrics}); accordingly, Tables~\ref{tab:exp_metrics_gcn_gat} and~\ref{tab:ra_mi_gcn_gat} should be read as pre$\to$post contrasts per dataset. Finally, we treat GRS as corroborative, qualitative evidence given surrogate variability---the surrogate tree shows only minor, non-decisive changes to its rule set relative to the structural shifts captured by $\mathrm{GED}\Delta$.

%\noindent\textbf{Discussion:}
%\paragraph{Discussion.}
%\subsection{Discussion}\label{sec:discussion}
\section{Discussion}\label{sec:discussion}
\noindent\textbf{RQ1 (Transparency).} Our verifier makes unlearning \emph{visible} primarily along two complementary axes: (i) \emph{attribution}, via HS/ESD pre$\to$post shifts on the affected region, and (ii) \emph{structure}, via the proxy graph edit distance $\mathrm{GED}\Delta$ on fixed $k$-hop ego networks. In addition, $\mathrm{RA}$ confirms the absence of attribution on the deleted nodes (expected for deletion‑based methods), while HS/ESD and $\mathrm{GED}\Delta$ provide the visible evidence that the neighborhood’s influence routes and local structure were also altered. We record GRS only as an \emph{auxiliary}, qualitative cue given surrogate variability. Read together—rather than in isolation—these artifacts form a human-readable audit trail of forgetting (Fig.~\ref{fig:fig1} and Tables~\ref{tab:exp_metrics_gcn_gat} and~\ref{tab:ra_mi_gcn_gat}); the within-dataset reading noted in Section~\ref{subsec:metrics} applies to attribution magnitudes. \noindent\textbf{RQ2 (Effectiveness).} We deem unlearning effective when two conditions, read jointly, are met: (i) targets no longer carry attribution ($\mathrm{RA}_{\text{post}}\!\approx\!0$ for deletion-based methods), and (ii) indirect influence routes in the $k$-hop neighborhood are disrupted (large $\mathrm{GED}\Delta$ together with substantive HS/ESD pre$\to$post shifts). The per-method outcomes that satisfy these criteria are summarized in Section~\ref{sec:results} and Tables~\ref{tab:exp_metrics_gcn_gat}.

The primary goal of this research is to build a transparent, method-agnostic verifier for GNN unlearning that produces human-readable, multi-channel evidence (attribution and structural, with an auxiliary rule view) alongside a complementary privacy signal. We report MI ROC--AUC as a complementary, graph-wide privacy indicator. Because MI aggregates over the full graph and is sensitive to calibration, its post-unlearning changes can be small or mixed in direction. Our localized explanation deltas ($\Delta$RA, HS/ESD, GED$\Delta$) therefore provide the primary, transparent evidence of forgetting; MI serves as an additional risk check (Section~\ref{sec:results}; Tables~\ref{tab:exp_metrics_gcn_gat} and~\ref{tab:ra_mi_gcn_gat}). Our verification focuses on explanation deltas (vanilla‑saliency) and fixed 2‑hop proxy graphs; HS/ESD are interpreted within‑dataset and GED as a relative indicator, with GraphChef rules used diagnostically. 
This favors transparency and comparability across five benchmarks and two backbones, while refraining from claims beyond this lens (e.g., other tasks or explainers).
Accordingly, we present our results as transparent within‑lens verification of forgetting signals—clear and verifiable now, and naturally compatible with complementary checks readers may value.

\section{Conclusion}
This study introduces an explainability-driven framework for verifying GNN unlearning under GDPR-like ``right-to-be-forgotten'' mandates, using five metrics---RA, HS, ESD, GED, and auxiliary GRS---to compare pre- and post-deletion snapshots. 
Across two backbones and five benchmarks, deletion‑based methods (Retrain, GNNDelete) eliminate direct attribution on the forgotten nodes ($\mathrm{RA}_{\text{post}} = 0$) and induce large edits in local proxy graphs (high GED$\Delta$); GraphEditor reduces but does not remove influence (smaller GED$\Delta$ with non‑zero $\Delta$RA), while IDEA shows little change. These localized attribution and structural deltas provide human‑readable, method‑agnostic evidence of forgetting for developers and regulators, and can be used alongside accuracy, membership‑inference risk, and certified guarantees; MI is reported only as a graph‑wide cross‑check. Our current scope is transductive node classification with a 5\%: future work will vary the forget‑set ratio, consider alternative explainers, and refine the auxiliary symbolic (GRS) view by improving surrogate fidelity and aligning rule‑level changes with attribution and proxy‑graph evidence.

\begin{acks}
This work was supported by the National Research Foundation of Korea (NRF) grant funded by the Korea government(MSIT) (RS-2025-02217071), and in part by the Institute of Information \& Communications Technology Planning \& Evaluation (lITP) grant funded by the Korea government (MSIT) (RS-2025-02305436, Development of Digital Innovative Element Technologies for Rapid Prediction of Potential Complex Disasters and Continuous Disaster Prevention).
\end{acks}

\newpage
\section{Ethical Considerations}
This study uses only publicly available graph benchmarks and does not involve collecting or processing personal data, human‑subjects research, or deployment to end users. Still, sharing model explanations or illustrative outputs may inadvertently reveal information about individuals or communities; we recommend reporting aggregate metrics, masking identifiers, and avoiding per‑instance releases. Data deletion can also shift outcomes across groups, so downstream users should monitor basic subgroup trends and document any collateral effects. Finally, results are intended for research evaluation rather than legal compliance; adhere to dataset licenses and institutional policies, limit access to fine‑grained outputs, and favor efficient, low‑footprint runs.

\bibliographystyle{ACM-Reference-Format}
\bibliography{references}
\end{document}